\documentclass[10pt,twocolumn,letterpaper]{article}

\usepackage{cvpr}
\usepackage{times}
\usepackage{epsfig}
\usepackage{graphicx}
\usepackage{amsmath}
\usepackage{amssymb}

\def\etal{\emph{et al}.}
\def\etc{\emph{etc}.}
\def\eg{\emph{e.g.}}
\def\ie{\emph{i.e.}}
\usepackage{booktabs} % toprule
\usepackage{array}  % ��������
\usepackage[tight,normalsize,sf,SF]{subfigure}
\usepackage{multirow}
\usepackage{url}

\def\V{\mathcal{V}}
\def\N{\mathcal{N}}
\def\X{\mathcal{X}}

\usepackage[breaklinks=true,bookmarks=false]{hyperref}

\cvprfinalcopy % *** Uncomment this line for the final submission

\def\cvprPaperID{****} % *** Enter the CVPR Paper ID here

\ifcvprfinal\fi
\begin{document}

%%%%%%%%% TITLE
\title{GIFT: A Real-time and Scalable 3D Shape Search Engine}

\author{Song Bai$^{1}$ \, Xiang Bai$^{1}$ \, Zhichao Zhou$^1$ \, Zhaoxiang Zhang$^2$ \, Longin Jan Latecki$^3$\\
$^1$Huazhong University of Science and Technology, $^3$Temple University \\
$^2$CAS Center for Excellence in Brain Science and Intelligence Technology, CASIA \\
{\tt\small \{songbai,xbai,zzc\}@hust.edu.cn} \,
{\tt\small zhaoxiang.zhang@ia.ac.cn}\, {\tt\small latecki@temple.edu}
}

\maketitle
\thispagestyle{empty}

%%%%%%%%% ABSTRACT
\begin{abstract}
\vspace{-2ex}
Projective analysis is an important solution for 3D shape retrieval, since human visual perceptions of 3D shapes rely on various 2D observations from different view points. Although multiple informative and discriminative views are utilized, most projection-based retrieval systems suffer from heavy computational cost, thus cannot satisfy the basic requirement of scalability for search engines.

In this paper, we present a real-time 3D shape search engine based on the projective images of 3D shapes.
The real-time property of our search engine results from the following aspects:
(1) efficient projection and view feature extraction using GPU acceleration;
(2) the first inverted file, referred as F-IF, is utilized to speed up the procedure of multi-view matching;
(3) the second inverted file (S-IF), which captures a local distribution of 3D shapes in the feature manifold, is adopted for efficient context-based re-ranking.
As a result, for each query the retrieval task can be finished within one second despite the necessary cost of IO overhead. We name the proposed 3D shape search engine, which combines \textbf{G}PU acceleration and \textbf{I}nverted~\textbf{F}ile (\textbf{T}wice), as~\textbf{GIFT}.
Besides its high efficiency, GIFT also outperforms the state-of-the-art methods significantly in retrieval accuracy on various shape benchmarks and competitions.
\end{abstract}

%%%%%%%%% BODY TEXT
\vspace{-3ex}
\section{Introduction}
3D shape retrieval is a fundamental issue in computer vision and pattern recognition. With the rapid development of large scale public 3D repositories,~\eg,~Google 3D Warehouse or TurboSquid, and large scale shape benchmarks,~\eg,~ModelNet~\cite{ModelNet}, SHape REtrieval Contest (SHREC)~\cite{SHREC14,SHREC2015}, the scalability of 3D shape retrieval algorithms becomes increasingly important for practical applications. However, \emph{efficiency} issue has been more or less ignored by previous works, though enormous efforts have been devoted to retrieval \emph{effectiveness}, that is to say, to design informative and discriminative features~\cite{kokkinos2012intrinsic,bronstein2010scale,IDSC,fang1,fang2,li2014persistence,litman2014supervised} to boost the retrieval accuracy. As suggested in~\cite{SHREC14}, plenty of these algorithms do not scale up to large 3D shape databases due to their high time complexity.

Meanwhile, owing to the fact that human visual perception of 3D shapes depends upon 2D observations, projective analysis~\cite{makadia2010spherical} has became a basic and inherent tool in 3D shape domain for a long time, with applications to segmentation~\cite{WangGWC0C13}, matching~\cite{PANORAMA}, reconstruction,~\etc. Specifically in 3D shape retrieval, projection-based methods demonstrate impressive performances. Especially in recent years, the success of planar image representation~\cite{BoW,VLAT,zheng2015query}, makes it easier to describe 3D models using depth or silhouette projections.

\begin{figure*}[tb]
\begin{center}
\includegraphics[width=0.8\linewidth]{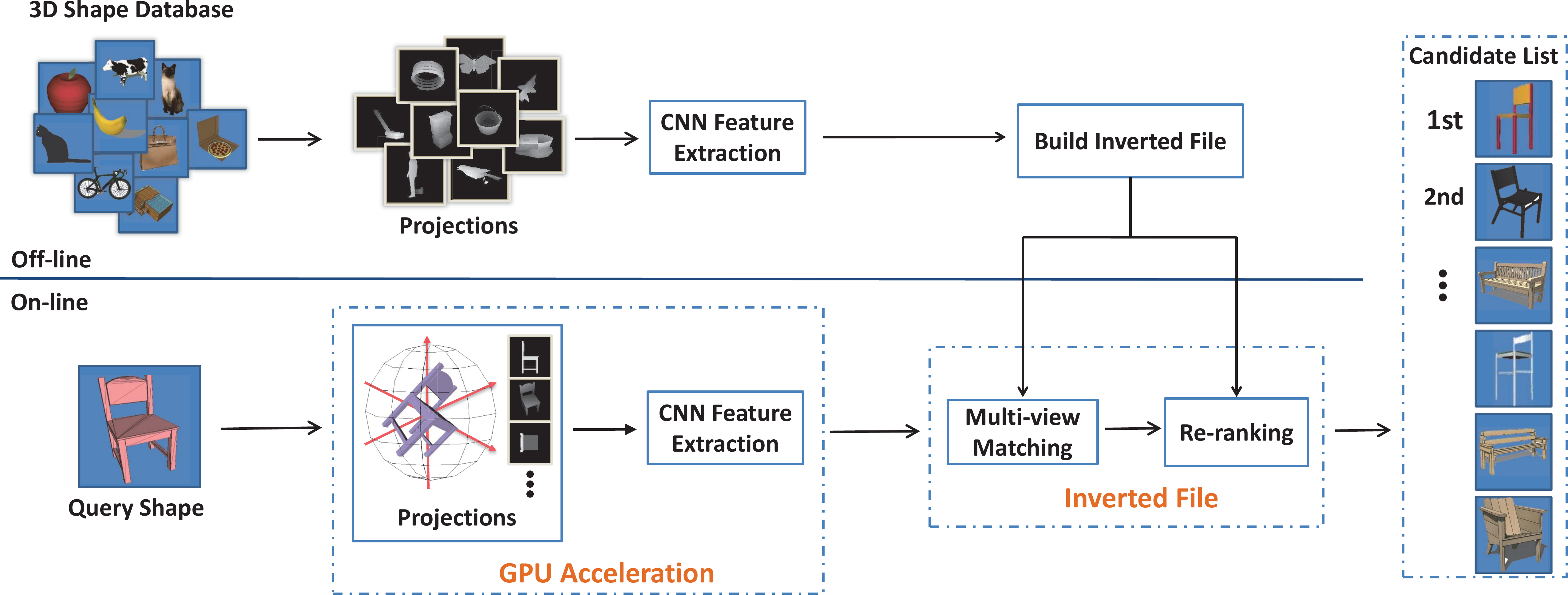}
\end{center}
\vspace{-2ex}
\caption{The structure of the proposed 3D shape search engine \textbf{GIFT}.}
\label{fig:pipeline}
\vspace{-2ex}
\end{figure*}

Generally, a typical 3D shape search engine is comprised of the following four components (see also Fig.~\ref{fig:pipeline}):
\begin{enumerate}
  \item \emph{Projection rendering.}~With a 3D model as input, the output of this component is a collection of projections. Most methods set an array of virtual cameras at pre-defined view points to capture views. These view points can be the vertices of a dodecahedron~\cite{LFD}, located on the unit sphere~\cite{VLAT}, or around the lateral surface of a cylinder~\cite{PANORAMA}. In most cases, pose normalization~\cite{NPCA} is needed for the sake of invariance to translation, rotation and scale changes.
  \item \emph{View feature extraction.}~The role of this component is to obtain multiple view representations, which affects the retrieval quality largely. A widely-used paradigm is Bag-of-Words (BoW)~\cite{BoW} model, since it has shown its superiority as natural image descriptors. However, in order to get better performances, many features~\cite{SHREC14} are of extremely high dimension. As a consequence, raw descriptor extraction (\eg,~SIFT~\cite{SIFT}), quantization and distance calculation are all time-consuming.
  \item \emph{Multi-view matching.}~This component establishes the correspondence between two sets of view features, and returns a matching cost between two 3D models. Since at least a set-to-set matching strategy~\cite{rodola2013efficient,rodola2014dense,rodola2013elastic,lian2013cm,havlena2014vocmatch} is required, this stage suffers from high time complexity even when using the simplest Hausdorff matching. Hence, the usage of algorithms incorporated with some more sophisticated matching strategies on large scale 3D datasets is limited due to their heavy computational cost.
  \item \emph{Re-ranking.}~It aims at refining the initial ranking list by using some extra information. For retrieval problems, since no prior or supervised information is available, contextual similarity measure is usually utilized. A classic context-based re-ranking methodology for shape retrieval is diffusion process~\cite{diffusion}, which exhibits outstanding performance on various datasets. However, as graph-based and iterative algorithms, many variants of diffusion process (\eg,~locally constrained diffusion process~\cite{LCDP}), generally require the computational complexity of $O(TN^3)$, where $N$ is the total number of shapes in the database and $T$ is the number of iterations. In this sense, diffusion process does not seem to be applicable for real-time analysis.
\end{enumerate}

In this paper, we present a real-time 3D shape search engine using projections that includes all the aforementioned components.
It combines \textbf{G}raphics Processing Unit (GPU) acceleration and \textbf{I}nverted~\textbf{F}ile (\textbf{T}wice), hence we name it~\textbf{GIFT}. In on-line processing, once a user submits a query shape, GIFT can react and present the retrieved shapes within one second (the off-line preprocessing operations, such as CNN model training and inverted file establishment, are excluded). GIFT is evaluated on several popular 3D benchmarks datasets, especially on one track of SHape REtrieval Contest (SHREC) which focuses on scalable 3D retrieval. The experimental results on retrieval accuracy and query time demonstrate the capability of GIFT in handling large scale data.

In summary, our main contributions are as follows.
Firstly, GPU is used to speed up the procedure of projection rendering and feature extraction. Secondly, in multi-view matching procedure, a robust version of Hausdorff distance for noise data is approximated with an inverted file, which allows for extremely efficient matching between two view sets without impairing the retrieval performances too much. Thirdly, in the re-ranking component, a new context-based algorithm based on fuzzy set theory is proposed. Different from diffusion processes of high time complexity, our re-ranking here is ultra time efficient on the account of using inverted file again.

%The rest of paper is organized as follows:
% \section{Related Work}
\section{Proposed Search Engine}
%In this section, a detailed introduction to each component of the proposed search engine is given.

\subsection{Projection Rendering} \label{sec:projections}
Prior to projection rendering, pose normalization for each 3D shape is needed in order to attain invariance to some common geometrical transformations. However, apart from many pervious algorithms~\cite{2D_3D_Hybrid,PANORAMA,NPCA} that require rotation normalization using some Principal Component Analysis (PCA) techniques, we only normalize the scale and the translation in our system. Our concerns are two-fold: 1) PCA techniques are not always stable, especially when dealing with some specific geometrical characteristics such as symmetries, large planar or bumpy surfaces; 2) the view feature used in our system can tolerate the rotation issue to a certain extent, though cannot be completely invariant to such changes. In fact, we observe that if enough projections (more than $25$ in our experiments) are used, one can achieve reliable performances.

%This allows us to address the problem that PCA techniques are not always stable, especially when dealing with specific geometrical characteristics such as symmetries, large planar or bumpy surfaces.
%The view features used in our system can tolerate rotation changes to a certain extent, though they are not completely invariant to such changes.

The projection procedure is as follows. Firstly, we place the centroid of each 3D shape at the origin of a spherical coordinate system, and resize the maximum polar distance of the points on the surface of the shape to unit length. Then $N_v$ virtual cameras are set on the unit sphere evenly, and they are located by the azimuth $\theta_{az}$ and the elevation $\theta_{el}$ angles. At last, we render one projected view in depth buffer at each combination of $\theta_{az}$ and $\theta_{el}$. For the sake of speed, GPU is utilized here such that for each 3D shape, the average time cost of rendering $64$ projections is only $30ms$.

\subsection{Feature Extraction via GPU Acceleration}
Feature design has been a crucial problem in 3D shape retrieval for a long time owing to its great influence on the retrieval accuracy. Though extensively studied, almost all the existing algorithms ignore the efficiency of the feature extraction.

To this end, our search engine adopts GPU to accelerate the procedure of feature extraction. Impressed by the superior performance of deep learning approaches in various visual tasks, we propose to use the activation of a Convolutional Neural Network (CNN). The CNN used here takes depth images as input, and the loss function is exerted on the classification error for projections. The network architecture consists of five successive convolutional layers and three fully connected layers as in~\cite{vgg}. We normalize each activation in its Euclidean norm to avoid scale changes. It only takes $56ms$ on average to extract the view features for a 3D model.

Since no prior information is available to judge the discriminative power of activations of different layers, we propose a robust re-ranking algorithm described in Sec.~\ref{sec:re_ranking}. It can fuse those homogenous features efficiently based on fuzzy set theory.

\subsection{Inverted File for Multi-view Matching}
Consider a query shape $x_q$ and a shape $x_p$ from the database $\X=\{x_1,x_2,\dots,x_N\}$. Let $\V$ denote a mapping function from 3D shapes to their feature sets. We can obtain two sets $\V(x_q)=\{q_1,q_2,\dots,q_{N_v}\}$ and $\V(x_p)=\{p_1,p_2,\dots,p_{N_v}\}$ respectively, where $N_v$ is the number of views. $q_i$ (or $p_i$) denotes the view feature assigned to the $i$-th view of shape $x_q$ (or $x_p$).

A 3D shape search engine requires a multi-view matching component to establish a correspondence between two sets of view features. These matching strategies are usually metrics defined on sets (\eg,~Hausdorff distance) or graph matching algorithms (\eg,~Hungarian method, Dynamic Programming, clock-matching). However, these pairwise strategies are time-consuming for a real-time search engine. Among them, Hausdorff distance may be the most efficient one, since it only requires some simple algebraic operations without sophisticated optimizations.

Recall that the standard Hausdorff distance measures the difference between two sets, and it is defined as
\begin{equation}
\label{eq:HD}
D(x_q, x_p)=\max_{q_i \in \V(x_q)}\min_{p_j \in \V(x_p)}d(q_i, p_j),
\end{equation}
where function $d(\cdot)$ measures the distance between two input vectors. In order to eliminate the disturbance of isolated views in the query view set, a more robust version of Hausdorff distance is given by
\begin{equation}
\label{eq:MHD}
D(x_q, x_p)=\frac1{N_v}\sum_{q_i \in \V(x_q)}\min_{p_j \in \V(x_p)}d(q_i, p_j).
\end{equation}
For the convenience of analysis, we consider its dual form in the similarity space as
\begin{equation}
\label{eq:MHS}
S(x_q, x_p)=\frac1{N_v}\sum_{q_i \in \V(x_q)}\max_{p_j \in \V(x_p)}s(q_i, p_j),
\end{equation}
where $s(\cdot)$ measures the similarity between the two input vectors. In this paper, we adopt the cosine similarity.

As can be seen from Eq.~\eqref{eq:MHD} and Eq.~\eqref{eq:MHS}, Hausdorff matching requires the time complexity $O(N\times {N_v}^2)$ for retrieving a given query (assuming that there are $N$ shapes in the database). Though the complexity grows linearly with respect to the database size, it is still intolerable when $N$ gets larger.
% in quadratic with respect to the database size, which is intolerable.
However, by analyzing Eq.~\eqref{eq:MHS}, we can make several observations: (1) let $s^\ast(q_i)=\max_{1\leq j\leq N_v}s(q_i, p_j)$, the similarity calculations of $s(q_i, p_j)$ are unnecessary when $s(q_i, p_j)<s^\ast(q_i)$, since these similarity values are unused due to the $max$ operation,~\ie,~only $s^\ast(q_i)$ is kept; (2) when considering from the query side, we can find that $s^\ast(q_i)$ counts little to the final matching cost if $s^\ast(q_i)<\xi$ and $\xi$ is a small threshold. Those observations suggest that although the matching function in Eq.~\eqref{eq:MHS} requires the calculation of all the pairwise similarities between two view sets, some similarity calculations, which generate small values, can be eliminated without impairing the retrieval performance too much.

In order to avoid these unnecessary operations and improve the efficiency of multi-view matching procedure, we adopt~\textbf{inverted file} to approximate Eq.~\eqref{eq:MHS} by adding the Kronecker delta response as
\begin{small}
\begin{equation}
\label{eq:AMHS}
S(x_q, x_p)=\frac1{N_v}\sum_{q_i \in \V(x_q)}\max_{p_j \in \V(x_p)}s(q_i, p_j)\cdot\delta_{c(q_i),c(p_j)},
\end{equation}
\end{small}
where $\delta_{x,y}=1$ if $x=y$, and $\delta_{x,y}=0$ if $x\neq y$.
The quantizer
$c(x)= \arg\min_{1\leq i\leq K}\|x-b_i\|^2$
maps the input feature into an integer index that corresponds to the nearest codeword of the given vocabulary $B=\{b_1,b_2,\dots,b_K\}$. As a result, the contribution of $p_j$, which satisfies $c(q_i)\neq c(p_j)$, to the similarity measure can be directly set to zero, without estimating $s(q_i, p_j)$ explicitly.
%Note that Eq.~\eqref{eq:AMHS} is only determined by the query side, so we can easily calculate the similarity values to the query on-the-fly.

\begin{figure}[tb]
\centering
\includegraphics[width=0.9\linewidth]{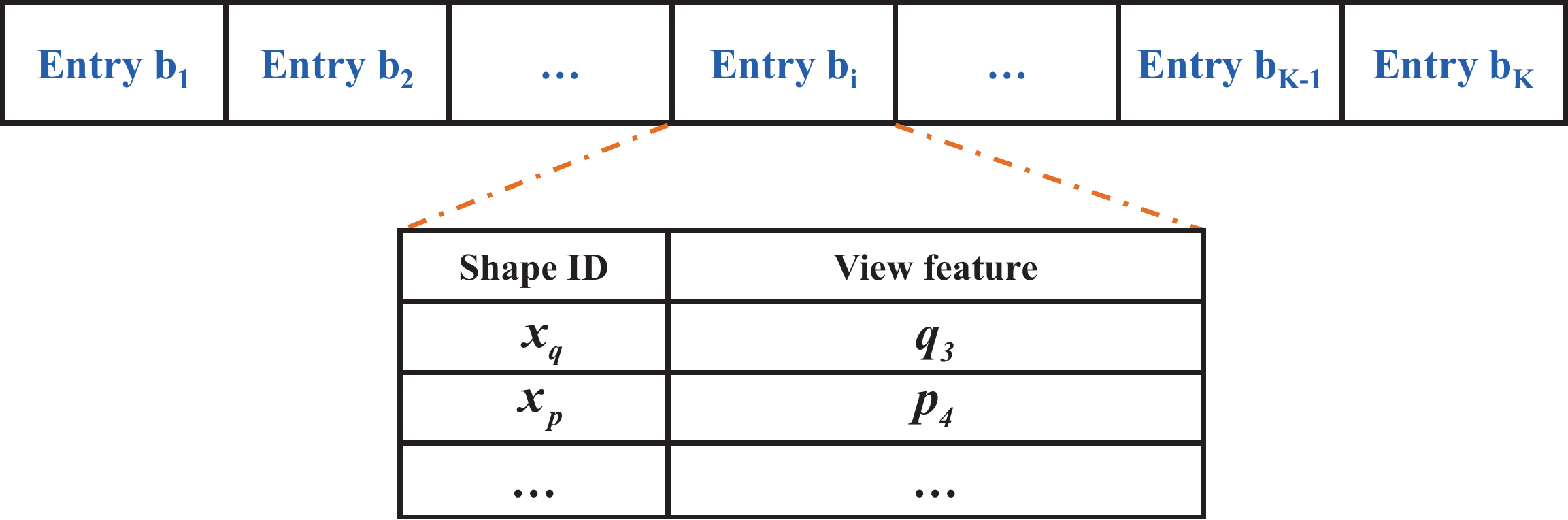}
\caption{The structure of the first inverted file.}
\label{fig:FIF}
\vspace{-2ex}
\end{figure}

In conclusion, our inverted file for multi-view matching is built as illustrated in Fig.~\ref{fig:FIF}. For each view feature, we store it and its corresponding shape ID in the nearest codeword. It should be mentioned that we can also use Multiple Assignment (MA),~\ie,~assign each view to multiple codewords, to improve the matching precision at the sacrifice of memory cost and on-line query time.
\subsection{Inverted File for Re-ranking} \label{sec:re_ranking}
A typical search engine usually involves a re-ranking component~\cite{reranking}, aiming at refining the initial candidate list by using some contextual information. In GIFT, we propose a new contextual similarity measure called Aggregated Contextual Activation (ACA), which follows the same principles as diffusion process~\cite{diffusion},~\ie,~the similarity between two shapes should go beyond their pairwise formulation and is influenced by their contextual distributions along the underlying data manifold. However, apart from diffusion process which has high time complexity, ACA enables real-time re-ranking, which can be applied to large scale data.

Let $\N_k(x_q)$ denote the neighbor set of $x_q$, which contains its top-$k$ neighbors. Similar to~\cite{zhang2015query}, our basic idea is that the similarity between two shapes can be more reliably measured by comparing their neighbors using Jaccard similarity as
\begin{equation} \label{equation:J}
S^{'}(x_q, x_p) = \frac{|\N_k(x_q) \cap \N_k(x_p)|}{|\N_k(x_q) \cup \N_k(x_p)|}.
\end{equation}
One can find that the neighbors are treated equally in Eq.~\eqref{equation:J}. However the top-ranked neighbors are more likely to be true positives. So a more proper behavior is increasing the weights of top-ranked neighbors.

To achieve this, we propose to define the neighbor set using fuzzy set theory. Different from classical (crisp) set theory where each element either belongs or does not belong to the set, fuzzy set theory allows a gradual assessment of the membership of elements in a set. We utilize $S(x_q, x_i)$ to measure the membership grade of $x_i$ in the neighbor set of $x_q$. Accordingly, Eq.~\eqref{equation:J} is re-written as

\begin{equation} \label{equation:J1}
S^{'}(x_q, x_p) = \frac{ \sum\limits_{x_i\in\N_k(x_q)\cap\N_k(x_p)}{\min\left(S(x_q, x_i),S(x_p, x_i)\right)} }{\sum\limits_{x_i\in\N_k(x_q)\bigcup\N_k(x_p)}{\max\left(S(x_q, x_i),S(x_p, x_i)\right)}}.
\end{equation}
Since considering equal-sized vector comparison is more convenient in real computational applications, we use $F\in \mathbb{R}^N$ to encode the membership values. The $i$-th element in $F_q$ is given as
\begin{equation} \label{equation:J2}
    F_q[i]=
   \begin{cases}
   S(x_q, x_i)~~~~~~~~~~&\mbox{if $x_i \in \N_k(x_q)$}\\
   0 &\mbox{otherwise}.
   \end{cases}
\end{equation}
Based on this definition we replace Eq.~\ref{equation:J1} with
\begin{equation} \label{equation:sca}
S^{'}(x_q, x_p) = \frac{\sum_{i = 1}^{N}\min\left(F_q[i], F_p[i]\right)}{\sum_{i = 1}^{N}\max\left(F_q[i], F_p[i]\right)}.
\end{equation}
Considering vector $F_q$ is sparse, we can view it as sparse activation of shape $x_q$, where the activation at coordinate $i$ is the membership grade of $x_i$ in the neighbor set $\N_k(x_q)$. Eq.~\eqref{equation:sca} utilizes the sparse activations $F_q$ and $F_p$ to define the new contextual shape similarity measure.

Note that all the above analysis is carried out for only one similarity measure. However, in our specific scenario, the outputs of different layers of CNN are usually at different abstraction resolutions.

For example, two different layers of CNN lead to two different similarities $S^{(1)}$ and $S^{(2)}$ by Eq.~\eqref{eq:MHS}, which in turn yield two different sparse activations
$F_q^{(1)}$ and $F_q^{(2)}$ by Eq.~\eqref{equation:J2}. Since no prior information is available to assess their discriminative power,
our goal now is
to fuse them in a unsupervised way.
%to distinguish different similarity measure, we add
For this we utilize the aggregation operation in fuzzy set theory, by which several fuzzy sets are combined in a desirable way to produce a single fuzzy set.
We consider two fuzzy sets represented by the
sparse activations $F_q^{(1)}$ and $F_q^{(2)}$
(the extension to more than two activations is similar) .
Their aggregation is then defined as
\begin{equation} \label{equation:ASA}
F_q = \left(\frac{{(F_q^{(1)}})^{\alpha}+{(F_q^{(2)}})^{\alpha}}2\right)^{\frac1\alpha},
\end{equation}
which computes the element-wise generalized means with exponent $\alpha$ of $F_q^{(1)}$ and $F_q^{(2)}$. Instead of using arithmetic mean, we use this generalized means ($\alpha$ is set to $0.5$ throughout our experiments). Our concern for this is to avoid the problem that some artificially large elements in $F_q$ dominate the similarity measure. This motivation is very similar to handling bursty visual elements in Bag-of-Words (BoW) model (see~\cite{jegou2009burstiness} for examples).

In summary, we call the feature in Eq.~\eqref{equation:ASA} Aggregated Contextual Activation (ACA). Next, we will introduce some improvements of Eq.~\eqref{equation:ASA} concerning its retrieval accuracy and computational efficiency.
%we can directly fuse them using the cartesian product of two neighbor sets to take the complementarity among them. Let $S^\alpha$ and $S^{\beta}$ denote two similarity measure obtained by two different layers in CNN, and two corresponding sparse activations $F_q^{\alpha}$ and $F_q^{\beta}$ can be generated for $x_q$ respectively.

%The cartesian product of $\N^{\alpha}_k(x_q)$ and $\N^{\beta}_k(x_q)$ is associated with the tensor product (or outer product) of $F_q^{\alpha}$ and $F_q^{\beta}$ as
%\begin{equation}
%F_q = F_q^{\alpha}\bigotimes F_q^{\beta}.
%\end{equation}
%The integrated sparse activation $F_q$, we term Tensor Sparse Activation (TSA), is vectorized later for re-ranking.
\vspace{-1ex}
\subsubsection{Improving Accuracy} \label{sec:IA}
\vspace{-1ex}
Similar to diffusion process, the proposed ACA requires an accurate estimation of the context in the data manifold. Here we provide two alternative ways to improve the retrieval performance of ACA without depriving its efficiency.

\vspace{1ex}\noindent\textbf{Neighbor Augmentation.}~The first one is to augment $F_q$ using the neighbors of second order,~\ie, the neighbors of the neighbors of $x_q$. Inspired by query expansion~\cite{PANORAMA}, the second order neighbors are added as
\begin{equation} \label{equation:LCE}
F^{(l)}_q := \frac{1}{|\N^{(l)}_k(x_q)|}\sum_{x_i\in \N^{(l)}_k(x_q)}{F^{(l)}_i}.
\end{equation}

\vspace{1ex}\noindent\textbf{Neighbor Co-augmentation.}~Our second improvement is to use a so-called ``neighbor co-augmentation". Specifically, the neighbors generated by one similarity measure are used to augment contextual activations of the other similarity measure, formally defined as
\begin{equation} \label{equation:co_LCE}
\begin{split}
F^{(1)}_q := \frac{1}{|\N^{(2)}_k(x_q)|}\sum_{x_i\in \N^{(2)}_k(x_q)}{F^{(1)}_i}, \\
F^{(2)}_q := \frac{1}{|\N^{(1)}_k(x_q)|}\sum_{x_i\in \N^{(1)}_k(x_q)}{F^{(2)}_i}.
\end{split}
\end{equation}
This formula is inspired by ``co-training"~\cite{co_t}. Essentially, one similarity measure tells the other one that ``I think these neighbors to be true positives, and lend them to you such that you can improve your own discriminative power".

Note that the size of neighbor set used here may be different from that used in Eq.~\eqref{equation:J2}. In order to distinguish them, we denote the size of neighbor set in Eq.~\eqref{equation:J2} as $k_1$, while that used in Eq.~\eqref{equation:LCE} and Eq.~\eqref{equation:co_LCE} as $k_2$.
\begin{figure}[tb]
\centering
\includegraphics[width=0.9\linewidth]{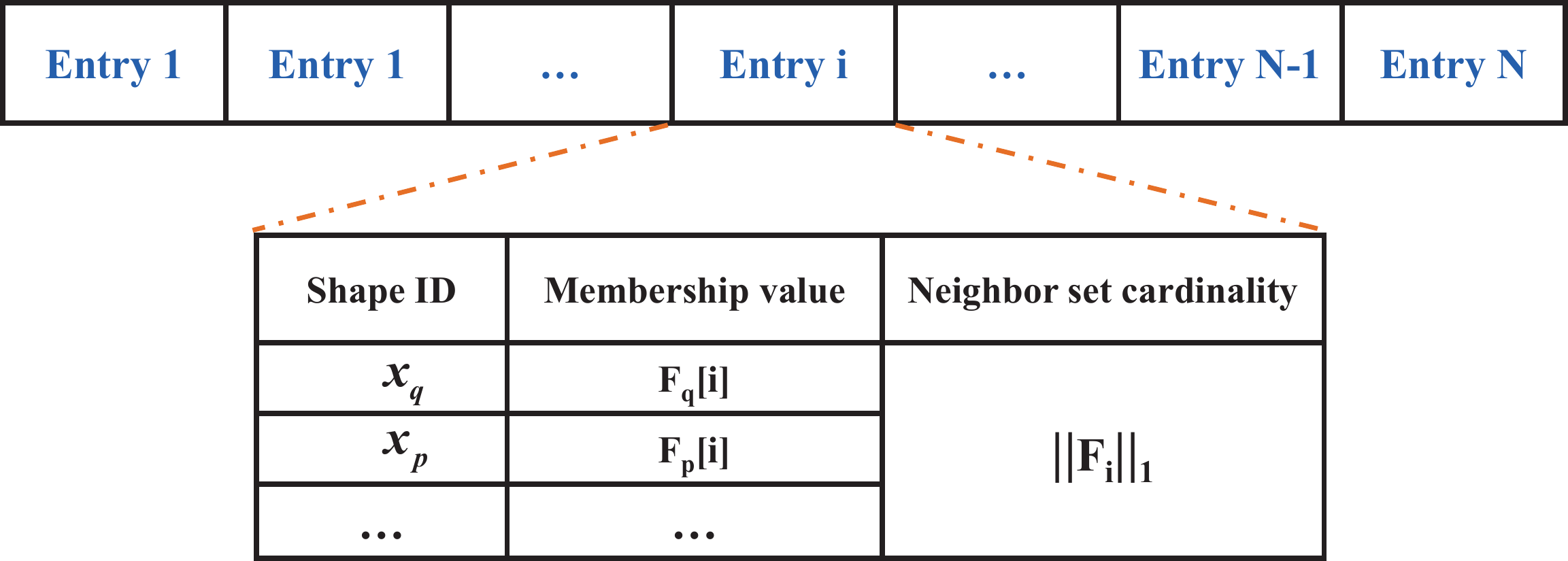}
\caption{The structure of the second inverted file.}
\label{fig:SIF}
\vspace{-3ex}
\end{figure}
%Since the second order neighbor are more likely to involve noise contextual information, an empirical rule for the determination of their values is $k_2<k_1$.
\vspace{-2ex}
\subsubsection{Improving Efficiency}
\vspace{-1ex}
Considering that the length of $F_q$ is $N$, one may doubt the efficiency of similarity computation in Eq.~\eqref{equation:sca}, especially when the database size $N$ is large. In fact, $F_q$ is a sparse vector, since $F_q$ only encodes the neighborhood structure of $x_q$, and the number of non-zero values is only determined by the size of $\N_k(x_q)$. This observation motivate us to utilize an \textbf{inverted file} again to leverage the sparsity of $F_q$.
%Different from the application of inverted file to Bag-of-Words (BoW)~\cite{BoF} representation in cosine similarity~\cite{NS},
Now we derive the feasibility of applying inverted file in Jaccard similarity theoretically.

The numerator in Eq.~\eqref{equation:sca} is computed as
\begin{equation} \label{equation:MIN_inverted}
\begin{split}
&\sum_i\min\left(F_q[i], F_p[i]\right)=\sum_{i|F_q[i]\neq0, F_p[i]\neq0}{\min(F_q[i],F_p[i])}  \\
&+\sum_{i|F_q[i]=0}{\min(F_q[i],F_p[i])}+\sum_{i|F_p[i]=0}{\min(F_q[i],F_p[i])}.
\end{split}
\end{equation}
Since all values of the aggregated contextual activation are non-negative, the last two items in Eq.~\eqref{equation:MIN_inverted} are equal to zero. Consequently, Eq.~\eqref{equation:MIN_inverted} can be simplified as
\begin{equation} \label{equation:MIN}
\sum_i\min\left(F_q[i], F_p[i]\right)=\sum_{i|F_q[i]\neq0, F_p[i]\neq0}{\min(F_q[i],F_p[i])},
\end{equation}
which only requires accessing non-zero entries of the query, and hence can be computed efficiently on-the-fly.

Although the calculation of the denominator in Eq.~\eqref{equation:sca} seems sophisticated,
%since it requires one to access all the vector components of $F_p[i^{\ast}]$, where $i^{\ast}=\arg\limits_{1\leq i\leq N}{F_q[i]=0}$. However,
it can be expressed as
\begin{equation} \label{equation:MAX}
\begin{split}
&\sum_{i}\max\left(F_q[i], F_p[i]\right) \\
&=\|F_q\|_1+\|F_p\|_1-\sum_{i}\min\left(F_q[i], F_p[i]\right) \\
&=\|F_q\|_1+\|F_p\|_1-\sum_{i|F_q[i]\neq0, F_p[i]\neq0}{\min(F_q[i],F_p[i])}.
\end{split}
\end{equation}
Besides the query-dependent operations (the first and the last items), Eq.~\eqref{equation:MAX} only involves an operation of $L_1$ norm calculation of $F_p$, which is simply equal to the cardinality of the fuzzy set $\N_k(x_p)$ and can be pre-computed off-line.

Our inverted file for re-ranking is built as illustrated in Fig.~\ref{fig:SIF}. It has exactly $N$ entries, and each entry corresponds to one shape in the database. For each entry, we first store the cardinality of its fuzzy neighbor set. Then, we find those shapes which have non-negative membership values in this entry. Those shape IDs and the membership values are stored in this entry.%It seems that the number of entries in this inverted file grows in quadratic with respect to the database size $N$, while the fact is that most entries are empty and can be eliminated since ....

%\vspace{2ex}\noindent\textbf{Ghost Points.}~Another way to produce such an accurate estimation is densifying the shape space using ghost points. Let $\{x_1,x_2,x_3\}$ denote a three point metric subspace, which is isometrically embedded into the plane $\mathbb{R}^2$ by an isometric embedding function $\kappa$. We can derive that the mean of $x_1$ and $x_2$ in the metric space is $\mu(x_1,x_2)=\kappa^{-1}\left(\frac12\kappa(x_1)+\frac12\kappa(x_2)\right)$. Consequently, the distance between an arbitrary point $x_3$ and $\mu(x_1,x_2)$ is
%\begin{equation} \label{eq:ghost}
%\begin{split}
%&~D\left(x_3, \mu(x_1,x_2)\right)^2 \\
%&=\kappa^{-1}\left( \|\kappa(x_3)-\kappa(\mu(x_1,x_2))\|^2 \right) \\
%&=\kappa^{-1}\left( \|\kappa(x_3)-\frac{\kappa(x_1)+\kappa(x_2)}2\|^2 \right) \\
%&=\resizebox{1\hsize}{!}{$\kappa^{-1}\left( \frac{\|\kappa(x_3)-\kappa(x_1)\|^2}{2}+\frac{\|\kappa(x_3)-\kappa(x_2)\|^2}{2}-\frac{\|\kappa(x_1)-\kappa(x_2)\|^2}{4} \right)$} \\
%&=\frac{D(x_3,x_1)}{2}^2+\frac{D(x_3,x_2)}{2}^2-\frac{D(x_1,x_2)}{4}^2.
%\end{split}
%\end{equation}
%Then $\mu(x_1,x_2)$, serving as ghost points, can be inserted by estimating its distance to $\forall x_3\in X$ according to Eq.~\eqref{eq:ghost}.
%
%In practice, one can add ghost point between any pair of points .

\section{Experiments}
In this section, we evaluate the performance of GIFT on different kinds of 3D shape retrieval tasks. The evaluation metrics used in this paper include mean average precision (MAP), area under curve (AUC), Nearest Neighbor (NN), First Tier (FT) and Second Tier (ST). Refer to~\cite{ModelNet,PSB} for their detailed definitions.
%A thorough discussion about the effect of different components in GIFT is given in Sec.~\ref{sec:dis}, and the analysis on execute time is summarized in Sec.~\ref{sec:time}.

If not specified, we adopt the following setup throughout our experiments. The projection rendered for each shape is $N_v = 64$. For multi-view matching procedure, the approximate Hausdorff matching defined in Eq.~\eqref{eq:AMHS} with an inverted file of $256$ entries is used. Multiple Assignment is set to $2$. We use two pairwise similarity measures, which are calculated using features from convolutional layer $L_5$ and fully-connected layer $L_7$ respectively. In re-ranking component, each similarity measure generates one sparse activation $F_q$ to capture the contextual information for the 3D shape $x_q$, and neighbor co-augmentation in Eq.~\eqref{equation:co_LCE} is used to produce $F^{(1)}_q$ and $F^{(2)}_q$. Finally, both $F^{(1)}_q$ and $F^{(2)}_q$ are integrated by \eqref{equation:ASA} with exponent $\alpha=0.5$.

\subsection{ModelNet}
ModelNet is a large-scale 3D CAD model dataset introduced by Wu~\etal~\cite{ModelNet} recently, which contains $151,128$ 3D CAD models divided into $660$ object categories. Two subsets are used for evaluation,~\ie, ModelNet40 and ModelNet10. The former one contains $12,311$ models, and the latter one contains $4,899$ models. We evaluate the performance of GIFT on both subsets and adopt the same training and test split as in~\cite{ModelNet}, namely randomly selecting $100$ unique models per category from the subset, in which $80$ models are used for training the CNN model and the rest for testing the retrieval performance.

For comparison, we collected all the retrieval results publicly available\footnote{\url{http://modelnet.cs.princeton.edu/}}. The chosen methods are (Spherical Harmonic) SPH~\cite{SPH}, (Light Field descriptor) LFD~\cite{LFD}, PANORAMA~\cite{PANORAMA}, 3D ShapeNet~\cite{ModelNet}, DeepPano~\cite{shi2015deeppano} and MVCNN~\cite{MVCNN}.
As Table~\ref{table_modelnet} shows, GIFT outperforms all the state-of-the-art methods remarkably.
We also present the performance of two baseline methods,~\ie,~feature $L_5$ or $L_7$ with exact Hausdorff matching. As can be seen, $L_7$ achieves a better performance than $L_5$, and GIFT leads to a significant improvement over $L_7$ of 5.82\% in AUC, 5.31\% in MAP for ModelNet40 dataset, and 3.32\% in AUC, 3.07\% in MAP for ModelNet10 dataset.
\begin{table}[tb]
\small
\centering
\begin{tabular}{lp{1.07cm}<{\centering}p{1.07cm}<{\centering}p{1.07cm}<{\centering}p{1.07cm}<{\centering}}
\toprule
\multirow{2}{*}{Methods} & \multicolumn{2}{c}{ModelNet40}   & \multicolumn{2}{c}{ModelNet10}  \\
\cmidrule(lr){2-3} \cmidrule(l){4-5}
    & AUC & MAP & AUC & MAP \\
\midrule
SPH~\cite{SPH} & 34.47\% & 33.26\% & 45.97\% & 44.05\% \\
LFD~\cite{LFD} & 42.04\% & 40.91\% & 51.70\% & 49.82\% \\
PANORAMA~\cite{PANORAMA} & 45.00\% & 46.13\% & 60.72\% & 60.32\% \\
ShapeNets~\cite{ModelNet} & 49.94\% & 49.23\% & 69.28\% & 68.26\% \\
DeepPano~\cite{shi2015deeppano} & 77.63\% & 76.81\% & 85.45\% & 84.18\% \\
MVCNN~\cite{MVCNN} & - & 78.90\% & - & - \\
\midrule
$L_5$ & 63.70\% & 63.07\% & 78.19\% & 77.25\% \\
$L_7$ & 77.28\% & 76.63\% & 89.03\% & 88.05\% \\
GIFT & \textbf{83.10\%} & \textbf{81.94\%} & \textbf{92.35\%} & \textbf{91.12\%} \\
\bottomrule
\end{tabular}
\caption{The performance comparison with state-of-the-art on ModelNet40 and ModelNet10.}
\label{table_modelnet}
\vspace{-2ex}
\end{table}

\begin{figure}[tb]
\centering
\subfigure[]
{
\begin{minipage}[tb]{0.22\textwidth}
\includegraphics[width = 1\textwidth]{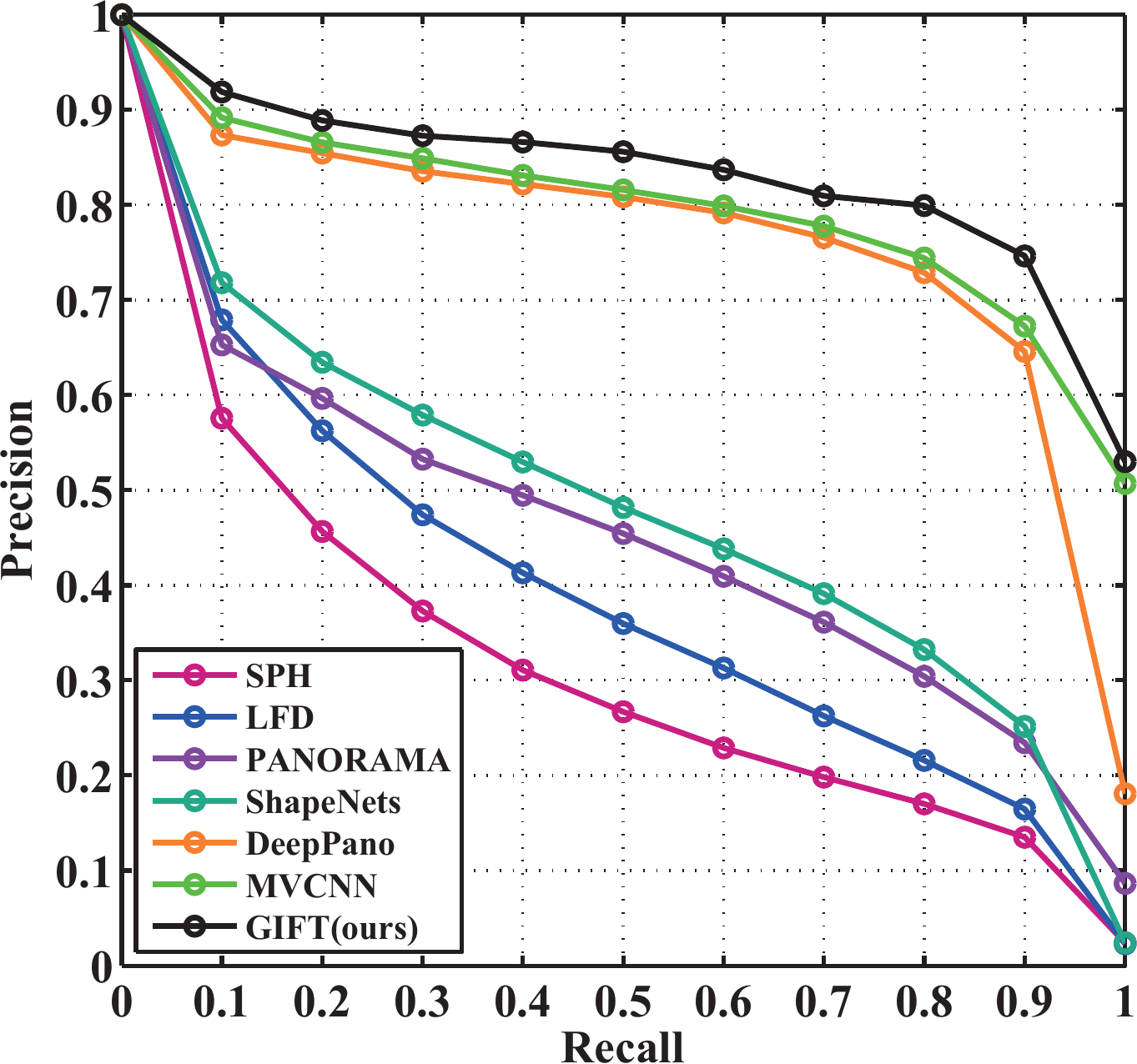}
\end{minipage}
\label{fig:fig:ModelNet_PR40}
}
\hspace{-2ex}
\subfigure[]
{
\begin{minipage}[tb]{0.22\textwidth}
\includegraphics[width = 1\textwidth]{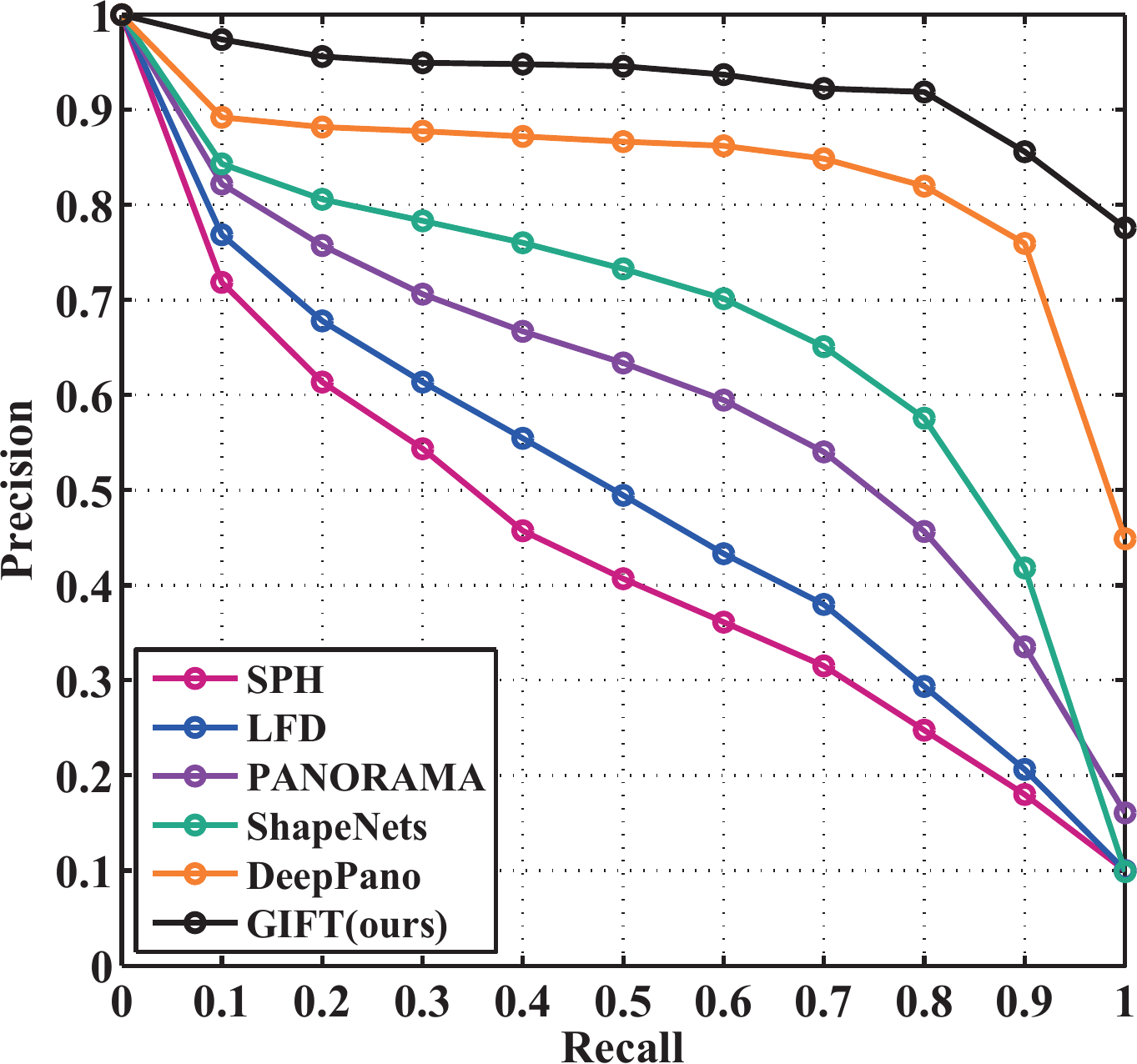}
\end{minipage}
\label{fig:ModelNet_PR10}
}
\caption{Precision-recall curves on ModelNet40 (a) and ModelNet10 (b).}
\label{fig:ModelNet_PR}
\vspace{-3ex}
\end{figure}

Fig.~\ref{fig:ModelNet_PR} compares the precision-recall curves.
It demonstrates again the discriminative power of the proposed search engine in 3D shape retrieval. Note that ModelNet also defines the 3D shape classification tasks. Considering GIFT is initially developed for real-time retrieval, its classification results are given in the supplementary material.

\vspace{-3ex}
\subsection{Large Scale Competition}
As the most authoritative 3D retrieval competition held each year, SHape REtrieval Contest (SHREC) pays much attention to the development of scalable algorithms gradually. Especially in recent years, several large scale tracks~\cite{SHREC2015}, such as SHREC14LSGTB~\cite{SHREC14}, are organized to test the scalability of algorithms. However, most algorithms that the participants submit are of high time complexity, and cannot be applied when the dataset becomes larger (millions or more). Here we choose SHREC14LSGTB dataset for a comprehensive evaluation. This dataset contains $8,987$ 3D models classified into $171$ classes, and each 3D shape is taken in turn as the query.
As for the feature extractor, we collected $54,728$ unrelated models from ModelNet~\cite{ModelNet} divided into $461$ categories to train a CNN model.

To keep the comparison fair, we choose two types of results from the survey paper~\cite{SHREC14} to present in Table~\ref{table_SHREC14}. The first type consists of the top-$5$ best-performing methods on retrieval accuracy, including PANORAMA~\cite{PANORAMA}, DBSVC, MR-BF-DSIFT, MR-D1SIFT and LCDR-DBSVC. The second type is the most efficient one,~\ie,~ZFDR~\cite{ZFDR}.
%Two Layer Coding with individual pair (TL-CODING)~\cite{bai20153d}

As can be seen from the table, excluding GIFT, the best performance is achieved by LCDR-DBSVC.
However, it requires $668.6s$ to return the retrieval results per query, which means that $69$ days are needed to finish the query task on the whole dataset. The reason behind such a high complexity lies in two aspects: 1) its visual feature is $270K$ dimensional, which is time consuming to compute, store and compare; 2) it adopts locally constrained diffusion process (LCDP)~\cite{LCDP} for re-ranking, while it is known that LCDP is an iterative graph-based algorithm of high time complexity. As for ZFDR, its average query time is shortened to $1.77s$ by computing parallel on $12$ cores. Unfortunately, ZFDR achieves much less accurate retrieval performance, and its FT is $13\%$ smaller than LCDR-DBSVC. In summary, a conclusion can be drawn that no method can achieve a good enough performance at a low time complexity.

By contrast, GIFT outperforms all these methods, including a very recent algorithm called Two Layer Coding (TLC)~\cite{bai20153d} which reports $0.585$ in FT. What is more important that GIFT can provide the retrieval results within $63.14ms$, which is $4$ orders of magnitude faster than LCDR-DBSVC.
Meanwhile, the two baseline methods $L_5$ and $L_7$ incur heavy query cost due to the usage of exact Hausdorff matching, which testifies the advantage of the proposed F-IF.
\begin{table}[tb]
\small
\centering
\begin{tabular}{lp{0.8cm}<{\centering}p{0.8cm}<{\centering}p{0.8cm}<{\centering}c}
\toprule
\multirow{2}{*}{Methods} & \multicolumn{3}{c}{Accuracy}   & \multirow{2}{*}{Query time}  \\
\cmidrule(lr){2-4}
 & NN & FT & ST & \\
\midrule
ZFDR            & 0.879 & 0.398 & 0.535 & 1.77$s$ \\
PANORAMA        & 0.859 & 0.436 & 0.560 & 370.2$s$ \\
DBSVC           & 0.868 & 0.438 & 0.563 & 62.66$s$ \\
%TL-CODING       & 0.879 & 0.456 & 0.585 & - \\
MR-BF-DSIFT     & 0.845 & 0.455 & 0.567 & 65.17$s$ \\
MR-D1SIFT       & 0.856 & 0.465 & 0.578 & 131.04$s$ \\
LCDR-DBSVC      & 0.864 & 0.528 & 0.661 & 668.6$s$ \\
\midrule
$L_5$	        & 0.879	& 0.460 & 0.592	& $22.73s$\\
$L_7$	        & 0.884	& 0.507	& 0.642	& $4.82s$\\
GIFT	        & \textbf{0.889}	& \textbf{0.567}	& \textbf{0.689} & $\textbf{63.14ms}$ \\
\bottomrule
\end{tabular}
\caption{The performance comparison on SHREC14LSGTB.}
\label{table_SHREC14}
\vspace{-2ex}
\end{table}

\subsection{Generic 3D Retrieval}
Following~\cite{Covariance}, we select three popular datasets for a generic evaluation, including Princeton Shape Benchmark (PSB)~\cite{PSB}, Watertight Models track of SHape REtrieval Contest 2007 (WM-SHREC07)~\cite{SHREC07} and McGill dataset~\cite{McGill}. Among them, PSB dataset is probably the first widely-used generic shape benchmark, and it consists of $907$ polygonal models divided into $92$ categories. WM-SHREC07 contains $400$ watertight models evenly distributed in $20$ classes, and is a representative competition held by SHREC community. McGill dataset focuses on non-rigid analysis, and contains $255$ articulated objects classified into $10$ classes. We train CNN on an independent TSB dataset~\cite{TSB}, and then use the trained CNN to extract view features for the shapes in all the three testing datasets.

In Table~\ref{table_generic}, a comprehensive comparison between GIFT and various state-the-art methods is presented, including LFD~\cite{LFD}, the curve-based method of Tabia~\etal~\cite{curve_analysis}, DESIRE descriptor~\cite{DESIRE}, total Bregman Divergences (tBD)~\cite{liu2012shape}, Covariance descriptor~\cite{Covariance}, the Hybrid of 2D and 3D descriptor~\cite{2D_3D_Hybrid}, Two Layer Coding (TLC)~\cite{bai20153d} and PANORAMA~\cite{PANORAMA}. As can be seen, GIFT exhibits encouraging discriminative ability in retrieval accuracy and achieves state-of-the-art performances consistently on all the three evaluation metrics.

\begin{table*}[tb]
\small
\centering
\begin{tabular}{lp{1.05cm}<{\centering}p{1.05cm}<{\centering}p{1.05cm}<{\centering}p{1.05cm}<{\centering}p{1.05cm}<{\centering}p{1.05cm}<{\centering}p{1.05cm}<{\centering}p{1.05cm}<{\centering}p{1.05cm}<{\centering}}
\toprule
\multirow{2}{*}{Methods} & \multicolumn{3}{c}{PSB dataset} & \multicolumn{3}{c}{WM-SHREC07 competition} & \multicolumn{3}{c}{McGill dataset} \\
\cmidrule(lr){2-4} \cmidrule(lr){5-7} \cmidrule(l){8-10}
                         &  NN    & FT     &  ST          &  NN    & FT     &  ST         &  NN    & FT     &  ST    \\
\midrule
LFD~\cite{LFD}                     & 0.657  & 0.380  & 0.487        & 0.923  & 0.526   & 0.662      & - & - & -               \\
Tabia~\etal~\cite{curve_analysis}  & -      & -      & -            & 0.853  & 0.527   & 0.639       & - & - & -           \\
DESIRE~\cite{DESIRE}               & 0.665  & 0.403   & 0.512       & 0.917  & 0.535   & 0.673      & - & - & -              \\
Makadia~\etal~\cite{makadia2010spherical}      & 0.673  & 0.412   & 0.502 & -      & -       & -           & - & - & -        \\
tBD~\cite{liu2012shape}            & 0.723  & -       & -           & -      & -       & -           & - & - & -        \\
Covariance~\cite{Covariance}       & -      & -      & -            & 0.930  & 0.623   & 0.737      & 0.977 & 0.732 & 0.818         \\
2D/3D Hybrid~\cite{2D_3D_Hybrid}   & 0.742  & 0.473   & 0.606       & 0.955  & 0.642   & 0.773       & 0.925 & 0.557 & 0.698          \\
PANORAMA~\cite{PANORAMA}           & 0.753  & 0.479   & 0.603       & 0.957  & 0.673   & 0.784      & 0.929 & 0.589 & 0.732      \\
%\hline
%Shape Vocabulary~\cite{raocong}    & 0.717  & 0.484   & 0.609 & 0.723     & - & - & - & - \\
PANORAMA + LRF~\cite{PANORAMA}     & 0.752  & 0.531   & 0.659       & 0.957  & 0.743  & 0.839    & 0.910 & 0.693 & 0.812         \\
TLC~\cite{bai20153d}         & 0.763  & 0.562   & 0.705       & 0.988  & 0.831   & 0.935       & 0.980 & 0.807 & 0.933          \\
\midrule
% 3DVFF~\cite{furuya2014fusing}      & -      & -       & -      & 0.841   & -      & -       & - & - \\
%\hline
$L_5$                          & 0.849   & 0.588   & 0.721   & 0.980	& 0.777	 &0.877		& 0.984 & 0.747 & 0.881    \\
$L_7$                          & 0.837  & 0.653    & 0.784   & 0.980 &	0.805 &	0.898	& 0.980 & 0.763 & 0.897  \\
GIFT                        & \textbf{0.849}& \textbf{0.712}& \textbf{0.830} & \textbf{0.990}& \textbf{0.949}& \textbf{0.990} & \textbf{0.984}& \textbf{0.905}& \textbf{0.973}   \\
\bottomrule
\end{tabular}
\vspace{1ex}
\caption{The performance comparison with other state-of-the-art algorithms on PSB dataset, WM-SHREC07 dataset and McGill dataset.}
\label{table_generic}
\vspace{-3ex}
\end{table*}

\vspace{-1ex}
\subsection{Execution Time} \label{sec:time}
\vspace{-1ex}
In addition to state-of-the-art performances on several datasets and competitions, the most important property of GIFT is the ``real-time" performance with the potential of handling large scale shape corpora. In Table~\ref{table_time_comp}, we give a deeper analysis of the time cost. The off-line operations mainly include projection rendering and feature extraction for database shapes, training CNN, and building two inverted files. As the table shows, the time cost of off-line operations varies significantly for different datasets. Among them, the most time-consuming operation is training CNN, followed by building the first inverted file with k-means. However, the average query time for different datasets can be controlled within one second, even for the biggest SHREC14LSGTB dataset.
\begin{table}[tb]
\small
\centering
\begin{tabular}{|l|p{2cm}<{\centering}|p{2.4cm}<{\centering}|}
\hline
Datasets & Off-line & On-line Indexing \\
\hline
\hline
ModelNet40      & $\approx0.7h$  & $27.02ms$ \\
ModelNet10      & $\approx0.3h$  & $10.25ms$ \\
\hline
SHREC14LSGTB    & $\approx8.5h$  & $63.14ms$ \\
\hline
PSB             &  \multirow{3}{*}{$\approx1.8h$}& $16.25ms$ \\
WMSHREC07       &  & $16.05ms$ \\
McGill          &  & $9.38ms$ \\
\hline
\end{tabular}
\caption{The time cost analysis of GIFT.}
\label{table_time_comp}
\vspace{-4ex}
\end{table}

\subsection{Parameter Discussion} \label{sec:dis}
Due to the space limitation, the discussion is conducted only on PSB dataset.
%First Tier is chosen to measure the retrieval performance.

\vspace{1ex}\noindent\textbf{Improvements Over Baseline.}~In Table~\ref{table_dis}, a thorough discussion is given about the influence of various components of GIFT. We can observe a consistent performance boost by those improvements. The performance jumps a lot especially when re-ranking component is embedded.
One should note a slight performance decrease when approximate Hausdorff matching with F-IF is used as compared with its exact version. However, as discussed below, the embedding with inverted file does not necessarily result in a poorer performance, but shortens the query time significantly.
\begin{table}[tb]
\small
\centering
\begin{tabular}{|l|c|p{1cm}<{\centering}|p{1cm}<{\centering}|p{1.2cm}<{\centering}|}
\hline
\multirow{2}{*}{Feature} & \multirow{2}{*}{Hausdorff}   & \multicolumn{2}{c|}{Re-ranking}   & \multirow{2}{*}{First Tier} \\
\cline{3-4}
 & &  $\alpha$ & NA & \\
\hline
\hline
$L_5$      & $\times$    &     &          & 0.588 \\
$L_7$     & $\times$    &     &          & 0.653 \\
\hline
\hline
$L_5+L_7$   & $\times$    &  1  &          & 0.688 \\
$L_5+L_7$   & $\times$    & 0.5 &          & 0.692 \\
$L_5+L_7$   & $\times$    & 0.5 & $\times$ & 0.710 \\
$L_5+L_7$  & $\times$    & 0.5 & $\surd$  & 0.717 \\
$L_5+L_7$   & $\surd$     & 0.5 & $\surd$  & 0.712 \\
\hline
\end{tabular}
\caption{The performance improvements brought by various components in GIFT over baseline. In column ``Hausdorff", $\surd$ denotes approximate Hausdorff matching in Eq.~\eqref{eq:AMHS}, while $\times$ denotes exact matching in Eq.~\eqref{eq:MHS}. Column ``$\alpha$" present the value of exponent in Eq.~\eqref{equation:ASA}. Column ``NA" describes the procedure of neighbor augmentation in Sec.~\ref{sec:IA}: $\surd$ is associated with Eq.~\eqref{equation:co_LCE} and $\times$ is associated with Eq.~\eqref{equation:LCE}. The blanks mean that this improvement is not used.}
%Column ``Ghost" indicates whether to insert ghost points.
\label{table_dis}
\vspace{-3ex}
\end{table}

\begin{figure}[tb]
\centering
\subfigure[]
{
\begin{minipage}[tb]{0.23\textwidth}
\includegraphics[width = 1\textwidth]{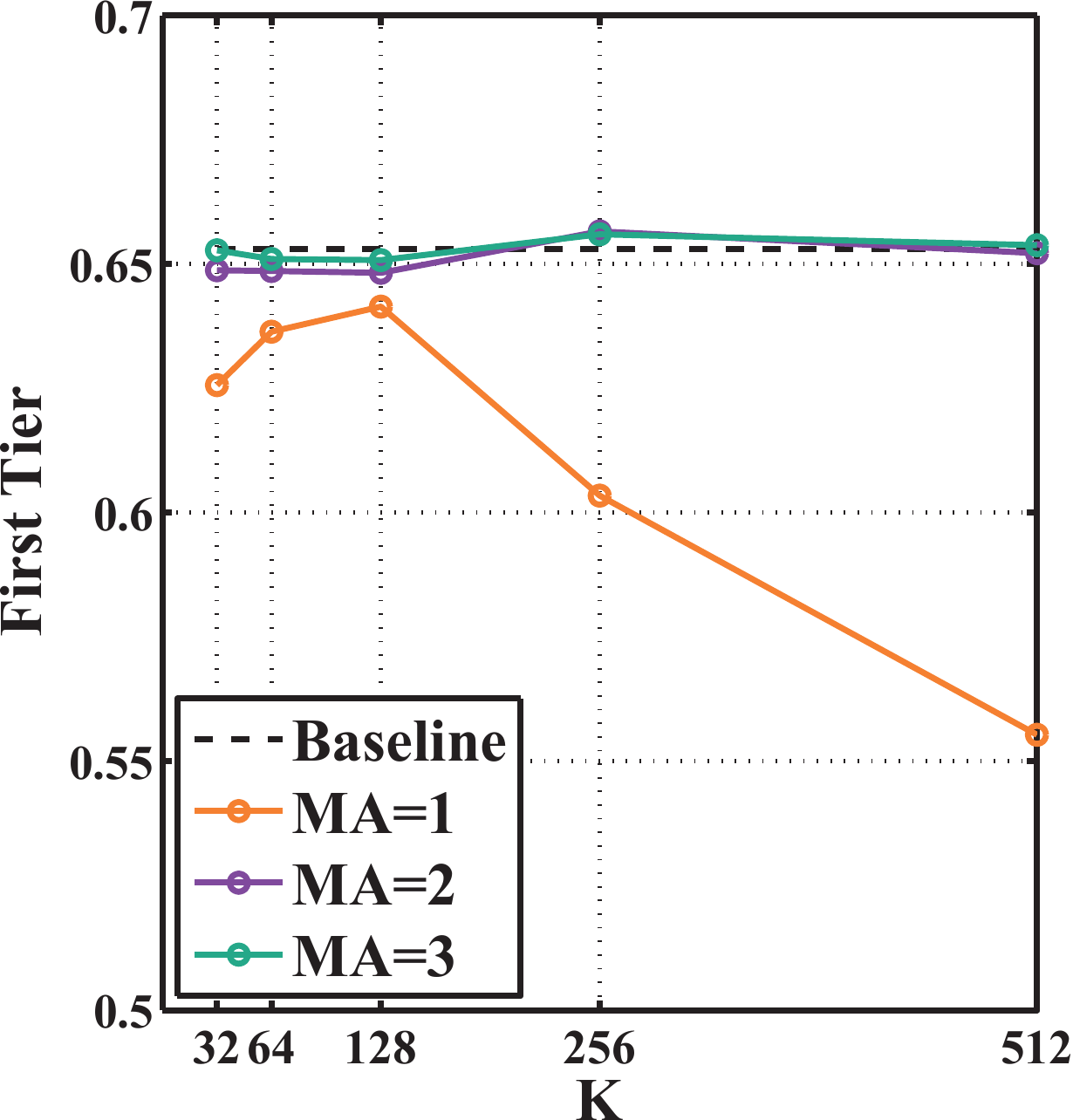}
\end{minipage}
\label{fig:size_FIF_accuracy}
}
\hspace{-2ex}
\subfigure[]
{
\begin{minipage}[tb]{0.23\textwidth}
\includegraphics[width = 1\textwidth, height = 26.5ex]{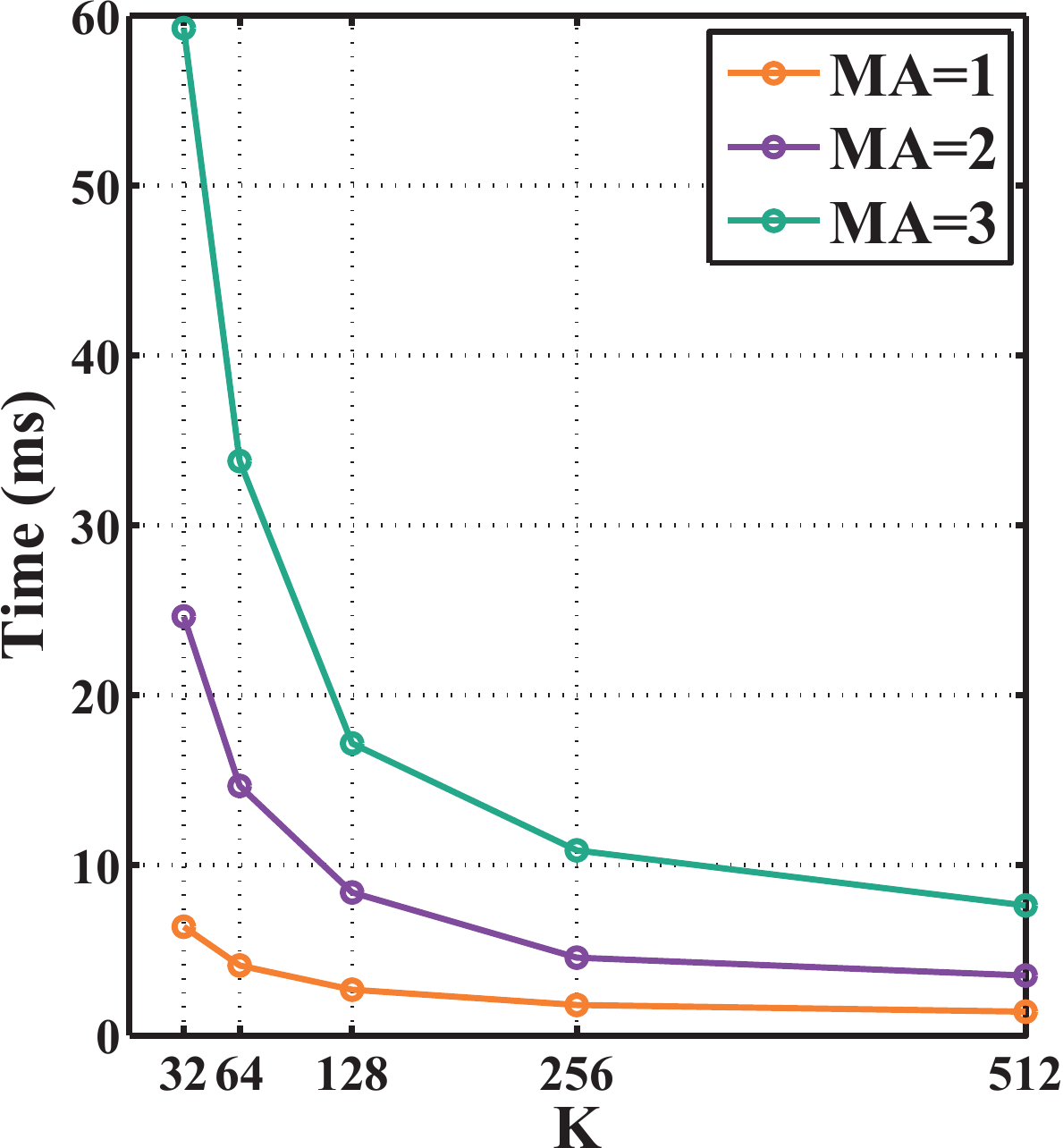}
\end{minipage}
\label{fig:size_FIF_time}
}
\caption{The performance difference between Hausdorff matching and its approximate version in terms of retrieval accuracy (a) and average query time (b).}
\label{fig:size_FIF}
\vspace{-3ex}
\end{figure}

\vspace{1ex}\noindent\textbf{Discussion on F-IF.}~In Fig.~\ref{fig:size_FIF}, we plot the retrieval performance and the average query time using feature $L_7$, as the number of entries used in the first inverted file changes. As Fig.~\ref{fig:size_FIF_accuracy} shows, the retrieval performance generally decreases with more entries, and multiple assignment can boost the retrieval performance significantly. However, it should be addressed that a better approximation to Eq.~\eqref{eq:MHS} using fewer entries (decreasing $K$) or larger multiple assignments (increasing MA) does not necessarily imply a better retrieval performance.
%see the fluctuation when MA$=2$ or $3$ and $K=256$.
For example, when $K=256$ and MA$=2$, the performance of approximate Hausdorff matching using inverted file surpasses the baseline using exact Hausdorff matching.
The reason for this ``abnormal" observation is that the principle of inverted file here is to reject those view matching operations that lead to smaller similarities, and sometimes they are noisy and false matching pairs which can be harmful to retrieval performance.

As can be seen from Fig.~\ref{fig:size_FIF_time}, the average query time is higher at smaller $K$ and larger MA, since the two cases both increase the number of candidate matchings in each entry. The baseline query time using exact Hausdorff matching is $0.69s$, which is at least one order of magnitude larger than the approximate one.

\begin{figure}[tb]
\centering
\includegraphics[width=0.95\linewidth]{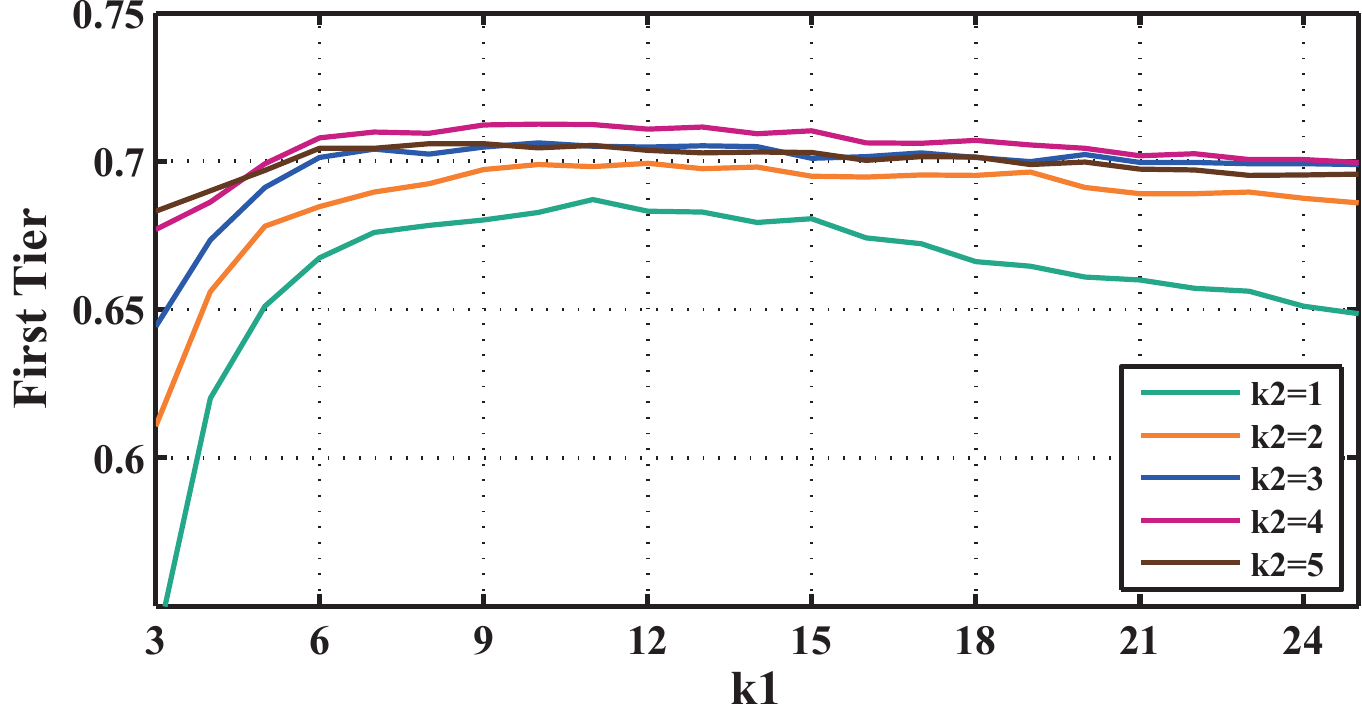}
\caption{The influence of neighbor set sizes $k1$ and $k2$ used in the second inverted file.}
\label{fig:k1k2}
\end{figure}

\vspace{1ex}\noindent\textbf{Discussion on S-IF.}~Two parameters,~$k_1$ and $k_2$, are involved in the second inverted file, which are determined empirically. We plot the influence of them in Fig.~\ref{fig:k1k2}. As can be drawn from the figure, when $k_1$ increases, the retrieval performance increases at first. Since noise contextual information can be included at a larger $k_1$, we can observe the performance decreases after $k_1>10$. Meanwhile, neighbor augmentation can boost the performance further. For example, the best performance is achieved when $k_2=4$. However, when $k_2=5$, the performance tends to decrease. One may find that the optimal value of $k_2$ is much smaller than that of $k_1$. The reason for this is that $k_2$ defines the size of the second order neighbor, which is more likely to return noise context compared with the first order neighbor defined by $k_1$.

\vspace{-1ex}
\section{Conclusions}
In the past years, 3D shape retrieval was evaluated with only small numbers of shapes. In this sense, the problem of 3D shape retrieval has stagnated for a long time. Only recently, shape community started to pay more attention to the scalable retrieval issue gradually. However, as suggested in~\cite{SHREC14}, most classical methods encounter severe obstacles when dealing with larger databases.

In this paper, we focus on the scalability of 3D shape retrieval algorithms, and build a well-designed 3D shape search engine called GIFT. In our retrieval system, GPU is utilized to accelerate the speed of projection rendering and view feature extraction, and two inverted files are embedded to enable real-time multi-view matching and re-ranking. As a result, the average query time is controlled within one second,
which clearly demonstrates the potential of GIFT for large scale 3D shape retrieval.
What is more impressive is that while preserving the high time efficiency, GIFT outperforms state-of-the-art methods in retrieval accuracy by a large margin. Therefore, we view the proposed search engine as a promising step towards larger 3D shape corpora.

We submitted a version of GIFT to the latest SHREC2016 large scale track (the results are available in~\url{https://shapenet.cs.stanford.edu/shrec16/}), and won the first place on perturbed dataset.

{\small
\bibliographystyle{ieee}
\bibliography{egbib}
}
\end{document}